\documentclass[10pt,twocolumn,letterpaper]{article}

\usepackage[pagenumbers]{cvpr} 

\usepackage{graphicx}
\usepackage{amsmath}
\usepackage{amssymb}
\usepackage{booktabs}
\usepackage{bm}
\usepackage{url}
\usepackage{comment}
\usepackage{color}

\usepackage[pagebackref,breaklinks,colorlinks]{hyperref}

\usepackage[capitalize]{cleveref}
\crefname{section}{Sec.}{Secs.}
\Crefname{section}{Section}{Sections}
\Crefname{table}{Table}{Tables}
\crefname{table}{Tab.}{Tabs.}

\def\cvprPaperID{6875} 
\def\confName{CVPR}
\def\confYear{2023}

\begin{document}

\title{Demystify Self-Attention in Vision Transformers from a Semantic Perspective: Analysis and Application}


\author{
  Leijie~Wu$^{1}$, 
  Song~Guo$^{1}$,
  Yaohong~Ding$^{1}$,  
  Junxiao~Wang$^{1}$,
  Wenchao~Xu$^{1}$,\\
  Richard Yida~Xu$^{2}$,
  and ~Jie~Zhang$^{1}$\\
  \textsuperscript{1}Department of Computing, The Hong Kong Polytechnic University\\
  \textsuperscript{2}Department of Mathematics, The Hong Kong Baptist University\\
  \texttt{\{lei-jie.wu,yaohong.ding,jieaa.zhang\}@connect.polyu.hk}\\
  \texttt{\{song.guo,junxiao.wang,wenchao.xu\}@polyu.edu.hk}, 
  \texttt{xuyida@hkbu.edu.hk} 
}

\maketitle

\begin{abstract}  
Self-attention mechanisms, especially multi-head self-attention (MSA), have achieved great success in many fields such as computer vision and natural language processing. 
However, many existing vision transformer (ViT) work simply inherent transformer designs from NLP to adapt vision tasks, while ignoring the fundamental difference between "how MSA works in image and language settings". 
Language naturally contains highly semantic structures that are directly interpretable by humans. Its basic unit (word) is discrete without redundant information, which readily supports interpretable studies on MSA mechanisms of language transformer.
In contrast, visual data exhibits a fundamentally different structure: Its basic unit (pixel) is a natural low-level representation with significant redundancies in the neighbourhood, which poses obvious challenges to the interpretability of MSA mechanism in ViT.
In this paper, we introduce a typical image processing technique, i.e., scale-invariant feature transforms (SIFTs), which maps low-level representations into mid-level spaces, and annotates extensive discrete keypoints with semantically rich information.
Next, we construct a weighted patch interrelation analysis based on SIFT keypoints to capture the attention patterns hidden in patches with different semantic concentration
Interestingly, we find this quantitative analysis is not only an effective complement to the interpretability of MSA mechanisms in ViT, but can also be applied to 1) spurious correlation discovery and “prompting” during model inference, 2) and guided model pre-training acceleration. 
Experimental results on both applications show significant advantages over baselines, demonstrating the efficacy of our method. 
\end{abstract}


\section{Introduction}
\label{sec:introduction}


\begin{figure}[t]
    \centering
    \includegraphics[width=0.47\textwidth]{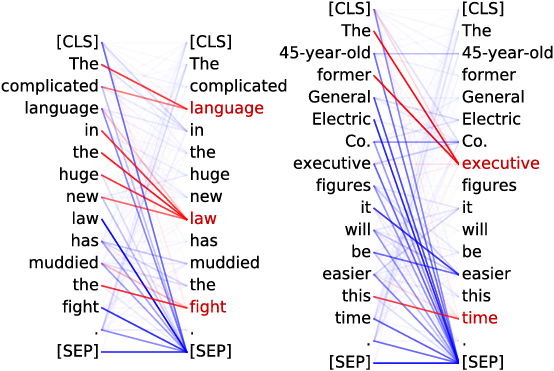}
    \caption{An example showing attention patterns in natural language correspond to human linguistic phenomena, where the line darkness represents the strength of attention weights. The attention pattern in this case is $94.3\%$ consistent with the grammar rule ``\textit{The Noun modifiers (e.g., determiners) will attend to their noun}'', which is colored {\color{red} red} in the Figure.}
    \label{fig:attention pattern in NLP}
\end{figure}

Towards robust representations for various downstream tasks, masked autoencoder frameworks such as BERT are widely used for model pre-training in the field of natural language processing (NLP) \cite{devlin2018bert}.
More specifically, they mask parts of the input sequence through Transformer-based encoders to generate latent representations from the rest, and then leverage decoders to predict the masked content.
This self-supervised pre-training approach leverages the multi-head self-attention (MSA) mechanism of the Transformer-based network architecture \cite{vaswani2017attention}.
As a landmark technique in feature modeling, MSA is now ubiquitous also in the field of computer vision.
The most widely accepted explanations for the success of MSAs are from the perspective of their weak inductive bias and capture of long-term dependencies \cite{dosovitskiy2021image,naseer2021intriguing,chu2021twins,mao2022towards}.
From a convolutional neural network perspective, MSA is a transformation of all feature map points, with large size and data-specific kernels.
Due to the simplicity and effectiveness of feature modeling, MSA is considered to be more expressive than channel attention in convolutional layers \cite{cordonnier2020relationship}.

Since the discrete basic units (i.e. words or tokens) abstracted by humans in language are highly semantic and information-dense, encoders can learn bidirectional token-level interrelationships through sophisticated language understanding (e.g., grammar) and generate information containing compact semantics.
%
%
Recent studies also showed that pre-trained language models have captured substantial linguistic \cite{hewitt2019structural,clark2019does} and relational knowledge \cite{petroni2019language,talmor2020olmpics,jiang2020can} through pre-training on large-scale text corpora.
These methods typically design ``fill-in-the-blank'' questions based on predefined relationships.
For example, a manually created question ``Bob Dylan was born in \_'' for the language model is to answer the ``birthplace'' relation of ``Bob Dylan''.
Furthermore, the highly semantic information of language provides favorable conditions for many works to perform interpretable analysis on the MSA mechanism of Transformer, which is directly understandable by humans, i.e., the workflow of attention heads fully conforms with the syntactic and language concepts of coreference defined by human language grammars \cite{clark2019does,li2020does}. 
Fig.~\ref{fig:attention pattern in NLP} shows an example to illustrate that attention patterns in NLP correspond to linguistic phenomena (the grammar rule of ``\textit{The Noun modifiers will attend to their noun}'').

Inheriting the \textit{mask-and-reconstruct} principle from NLP, He et al. first extend the masked autoencodeing approach to the pre-training of Vision Transformer (ViT) model, which has gained great success on both model pre-training and inference \cite{he2022masked}. 
Many subsequent studies focus on altering designs directly on the basic structure from NLP to achieve good performance \cite{zhou2021ibot,wei2022masked,yuan2022volo},
while ignoring the fundamental difference in how the MSA mechanism works between image and language settings.
%
%
Image data, in contrast, is a natural signal whose basic units (i.e., pixels) contain only low-level representations and often exhibit severe spacial redundancy due to the image continuity in the neighbourhood,
%
%
which poses a significant challenge to the interpretable analysis of MSA mechanisms in ViT. 
As yet, only a few studies try to interpret how MSA mechanism works in ViT, but their explanations are all based on empirical observations about attention distributions \cite{ma2022visualizing} or model output \cite{amir2021deep} on low-level space, while interpretable semantic analysis or theoretical support are absent from literature.

In this paper, to quantify the semantic information in low-level image space, 
we introduce a traditional image processing techniques, called Scale-Invariant Feature Transform (SIFT)\cite{lowe2004distinctive}.
%
%
More specifically, SIFT first maps the low-level redundant image space into mid-level feature space with highly semantic information. 
In the new space, it extract rich semantic information from image pixels to automatically annotates various discrete feature keypoints.These keypoints are theoretically proved invariant in different situations (e.g., rotation, scaling, brightness, and orientation), which also naturally correspond to the principles of image understanding in human logic \cite{morel2011sift}. 
Based on semantic-rich SIFT keypoints, we establish a weighted patch interrelation measurement to capture hidden attention patterns of all heads. 
The dynamic mathematical statistics from massive results help us to capture the existence of attention bias (i.e., patches with high SIFT keypoint concentration will pay more attention to each other, while relatively ignore other low concentration patches).
We further conduct comprehensive analysis yo derive a summary of three different stages, which answers "\textit{Yes}" to the question: ``\textit{Is there a semantically-similar interrelation between patches with different SIFT keypoint concentration?}''. 
%

More interestingly, we find our quantitative analysis is not only an effective complement to the interpretability of existing MSA mechanisms, but can also be used for a variety of applications, including 1) discovering spurious correlations and prompting input during model inference to improve performance, 2) guiding model pre-training to speed up convergence.
The main contributions of this paper are summarized as follows:
\begin{itemize}
    \item We introduce SIFT to map the low-level redundant image space into the mid-level semantic feature space. It can automatically annotate invariant feature keypoints, and provide theoretical guarantee for measuring the semantic information level of different patches.
    
    \item We design a weighted quantitative analysis based on SIFT keypoints, which interprets that the attention patterns hidden in MSA mainly exploits the semantically-similar interrelation between patches with different SIFT keypoint concentration during inference.
    
    \item We further derive a range of well-suited applications from the interpretable analysis, including 1) discovering spurious correlations and prompting input during model inference to improve performance, and 2) guiding model pre-training to accelerate convergence.
\end{itemize}





\section{Key Elements Hidden in Image Patch}
\label{sec:The Key Elements for Image Recovery}

In this section, through the pre-experiments on image recovery, we reveal the fact that patches have significant differences on semantic information level with hidden interrelation. Then, we use SIFT to quantify the hidden semantic information level of different patches.

% ...

\subsection{Patch Interrelation in Image Recovery}
\label{Patch Interrelation in Image Recovery}

\begin{figure}[t]
    \centering
    \includegraphics[width=0.47\textwidth]{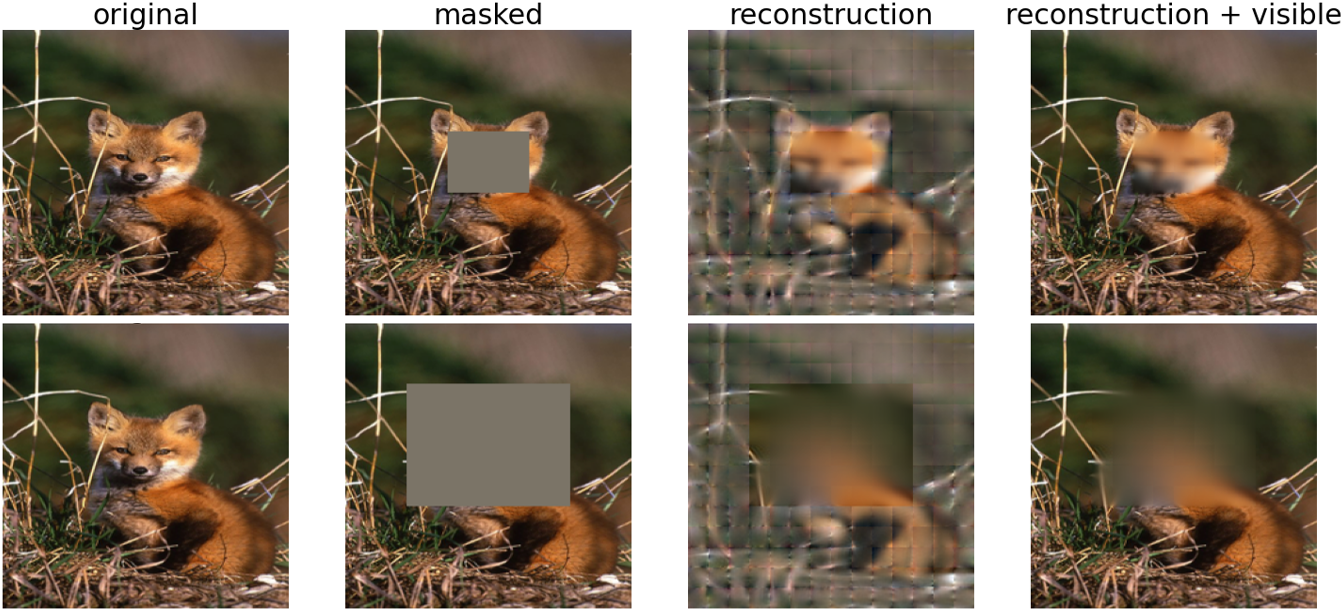}
    \caption{Image restoration results after adding different masks. We use the same pretrained ViT-B/16 model for image restoration, with original images from ILSVRC2012 (ImageNet 1K).}
    \label{fig:image recovery on different patches}
\end{figure}

As shown in the second ``\textit{masked}'' column of Fig.~\ref{fig:image recovery on different patches}, we add masks with different sizes to the original image and restore the masked image using the same pretrained ViT-B/16 model, where original images are from ILSVRC2012 (ImageNet 1K) .
When we only mask the fox face regions in the top row, model can extract enough interrelation information from remaining patches to recover the masked face regions.
However, if we extend the mask to the entire head area in the bottom row, model is unable to reconstruct the masked area from the remaining patches.

The above comparison results indicate the significant differences between patches and the absence of a few key patches will lead to model reconstruction capability degradation.
%
According to previous studies \cite{dosovitskiy2021image,he2022masked}, the key elements hidden in patches are different semantic information they contain, i.e., key features.
In addition, there is also a high degree of interrelation inference between patches containing key features, which together can provide sufficient semantic information to help the model understand parts, objects, and scenes in masked regions.
%
%
For example, in the first row of Fig.~\ref{fig:image recovery on different patches}, the patch with fox ears indicates that the masked area is likely to be a fox face.
When patches with key features are masked (see bottom row of Fig.~\ref{fig:image recovery on different patches}), the model cannot extract enough semantic information from the remaining patch interrelations to understand and recover the masked regions.
%

\subsection{SIFT Keypoint}
\label{SIFT Keypoint}

Before analyzing and quantifying the patch interrelation, we first need to identify the concentration of semantic information (i.e. key features) contained in different patches. %
As some mid-level representations with rich semantic information,
key features should be invariant in different situations, such as rotation, scale, brightness, and orientation.
Therefore, we introduce the classical scale-invariant feature transform (SIFT) to automatically annotate broad feature keypoints hidden in images \cite{lowe2004distinctive}.

\begin{figure}[t]
    \centering
    \includegraphics[width=0.48\textwidth]{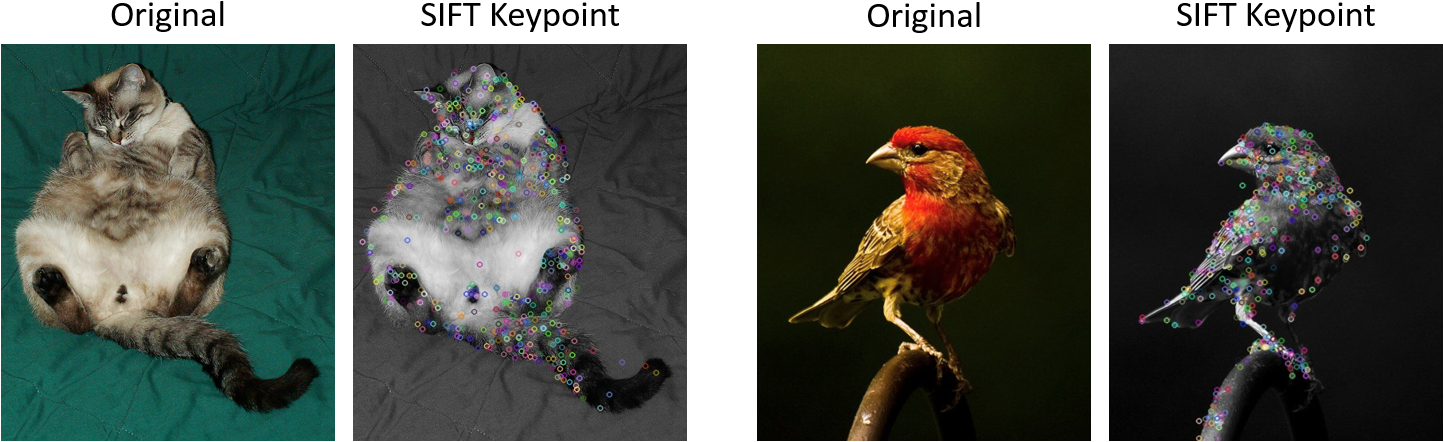}
    \caption{Illustration of SIFT keypoint annotations in an image. SIFT converts the original color image to a grayscale image, avoiding the effect of color on keypoint recognition.
    }
    \label{fig:sift keypoint}
\end{figure}

More specifically, SIFT first uses difference of Gaussian (DoG) to generate a set of results obtained from the same image at different scales.
%
%
Then, SIFT will calculate the directed gradient of each point and compare it with its neighbors to detect those local extreme points. 
Only those local extreme points that are invariant in different scale spaces are selected as keypoints.
A demonstration of SIFT keypoint annotation is shown in Fig.~\ref{fig:sift keypoint}, and we can observe that most SIFT keypoints are concentrated on the target `cat', which is highly consistent with our definition of key features in human semantic understanding.
%

With the help of SIFT keypoints, we can quantify the concentration of semantic information (i.e. key features) contained in different patches by counting the number of SIFT keypoints in each patch.
This quantitative metric will be used in the next section to reveal the patch interrelations hidden in the above image restoration pre-experiments.

\section{Patch Interrelation Analysis}
\label{sec:Visualization of Patch Interrelation}

In this section, we first introduce the background of ViT and its multi-head self-attention (MSA) mechanism, and then describe our SIFT keypoint-based measure that can reveal the patch interrelations during model inference.

\subsection{Preliminaries}
\label{subsec:Preliminaries}

For an original input image of resolution $(H, W)$, given that the resolution of each patch is $(P, P)$, ViT will split the image into a set of non-overlapping patches $[p_1, \cdots, p_N]$, where the patch number is $N=HW/P^2$.
By linear projection, these patches will be transformed into patch embeddings $[h_1, \cdots, h_N]$ with fixed vector size.
The patch embeddings are then fed into a series of stacked transformers, each containing a multi-head self-attention (MSA) mechanism with $L$ parallel SA heads.
More specifically, in an SA head, each patch embedding $h_i$ will be mapped to the query vector $q_i$, key vector $k_i$ and value vector $v_i$ vector by a separate linear projection.
The head will compute the attention weights $\alpha$ between all patch pairs by the softmax normalized dot product between the query and the key vector. 
The output $o$ of the head is a weighted sum of all value vectors.

\begin{equation}
\label{eq:compute attention}
    \alpha_{i,j} = \frac{\exp(q_i^Tk_j)}{\sum_{l=1}^n\exp(q_i^T,k_l)}, \quad o_i = \sum_{j=1}^n \alpha_{ij}v_j.
\end{equation}

When generating the next representation of the current patch $i$, the attention weight $\alpha_{ij}$ can be regarded as the `\textit{importance}' of every other patch $j$ to it, which is implicitly consistent with our patch interrelation.
Therefore, our measurement method is based on the attention weight distribution of each MSA transformer block.

We use the \textbf{ViT-Large} model (ViT-L/16) pre-trained on ImageNet-21K as our backbone, which has $24$ stacked transformer layers, each containing $16$ attention heads.
% ...

\subsection{Measurement of Patch Interrelation}
\label{subsec:Measurement of Patch Interrelation}

Traditional attention mechanisms follow human logic on image processing to improve model performance by focusing attention on some key regions (such as objects in images) \cite{hu2018squeeze,woo2018cbam}.
They usually use spatial or channel attention to individually highlight some key features, while ignoring the semantic understanding hidden in patch interrelations is also important during model inference.
Unlike traditional attention mechanisms, the computation process of the SA mechanism naturally involves comparisons between all possible patch pair combinations.

\begin{figure}[t]
  \centering
  \begin{subfigure}{0.54\linewidth}
    \includegraphics[width=\linewidth]{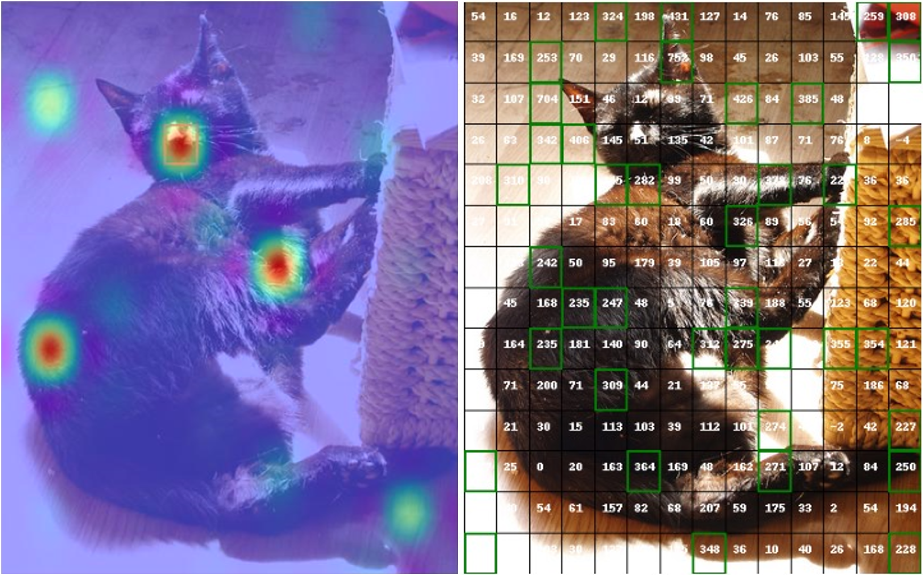}
    \caption{Attention value matrix and its visualization heatmap}
    \label{fig:attention heatmap with value}
  \end{subfigure}
  \begin{subfigure}{0.44\linewidth}
    \includegraphics[width=\linewidth]{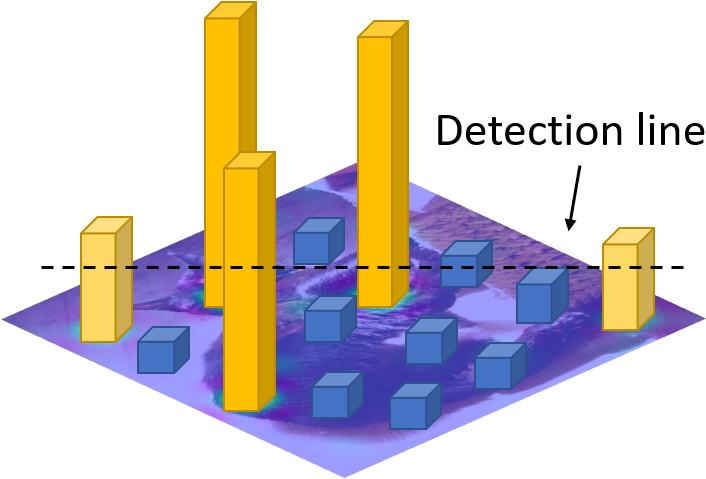}
    \caption{3D attention mountain-peak map visualization}
    \label{fig:3D attention visualization}
  \end{subfigure}
  \caption{The attention matrix $\bm{\alpha}_i$ of the patch $i$ computed on an SA head and its visualization. (zoom in to see detailed values within the matrix)}
  \label{fig:short}
  \vspace{-10pt}
\end{figure}

For example, when generating the next representation for patch $i$ in equation (\ref{eq:compute attention}), each SA head will compute a $16\times 16$ attention matrix $\bm{\alpha}_i = [\alpha_{ij}]$.
The value of $\alpha_{ij}$ in each patch $j\in\{N\}$ represents the importance of other patches $j$ to the patch $i$.
In other words, $\alpha_{ij}$ also means the strength of patch interrelation between patches $i$ and $j$.
The attention heatmap visualized from this matrix is shown in Fig.~\ref{fig:attention heatmap with value}.
The heatmap can also be transformed into a mountain-peak map by the 3D visualization in Fig.~\ref{fig:3D attention visualization}, where the mountain height on each patch is $\alpha_{ij}$.

To quantify patch interrelationships between patch $i$ and another patch $j$, as shown in Fig.~\ref{fig:3D attention visualization}, we design a detection line and define its height according to:
\begin{equation}
\label{eq:detection line}
    \Bar{H} = \gamma * \frac{\sum_{j=1}^N \alpha_{ij}}{N},
\end{equation}
Where $\gamma$ is an adjustment parameter that controls the height of the detection line.
Only patches $j$ with value $\alpha_{ij}$ higher than $\Bar{H}$ ($\alpha_{ij}\ge\Bar{H}$) will be filtered out as attended patches (see yellow bars in Fig.~\ref{fig:3D attention visualization}), the set of attended patches is denoted by $\mathcal{S}_i$.

\textbf{Weighted Patch Interrelation Analysis.} 
As we know from Sec.~\ref{sec:The Key Elements for Image Recovery}, patches are highly discriminative due to the set of semantic information they contain, i.e., the number of SIFT keypoints in the patch.
The set of attended patches $\mathcal{S}_i$ can thus be separated into two non-overlapping subsets: $\mathcal{S}_i^{\textit{Non}}$ includes attended patches that do not contain SIFT keypoints and $\mathcal{ S }_i^{\textit{Key}}$ includes attended patches that contain SIFT keypoints that satisfy $\mathcal{S}_i = \mathcal{S}_i^{\textit{Non}} \cup \mathcal{S}_i^{\textit{Key}}$ and $\mathcal{S}_i^{\textit{Non}} \cap \mathcal{S}_i^{\textit{Key}} = \emptyset$.
Intuitively, calculating the percentage of $\mathcal{S}_i^{\textit{Key}}$ in the entire attended patch set $\mathcal{S}_i$ can reflect the patch interrelation distribution (i.e., starting with patch $i$, whether the model is biased towards those patches contain SIFT keypoints among all attended patches).

However, it is not fair to treat each patch equally, as the SIFT keypoints they contain may vary widely, and patches containing more SIFT keypoints should have higher weights in our patch interrelation analysis.
Denote the number of SIFT keypoints in each patch $j\in \{N\}$ by $t_j$.
The weighted patch correlation analysis generated for patch $i$ on an SA head translates to:
\begin{equation}
\label{eq:weighted patch interrelation}
     \theta_i = \underbrace{ \frac{|\mathcal{S}_i^{\textit{Key}}|}{|\mathcal{S}_i^{\textit{Non}}| + |\mathcal{S}_i^{\textit{Key}}|} }_{\text{Unweighted}} \Rightarrow
     \theta_i = \underbrace{ \frac{ \sum\limits_{j\in\mathcal{S}_i^{\textit{Key}}} t_j }{ |\mathcal{S}_i^{\textit{Non}}| + \sum\limits_{j\in\mathcal{S}_i^{\textit{Key}}} t_j }  }_{\text{Weighted}}
\end{equation}

It is worth noting that each SA head generates a total $N$ attention matrix $\bm{\alpha}_i$ for different starting patch $i\in\{N\}$, and the starting patch $i$ can vary significantly due to the contained SIFT keypoints. 
They are divided into two non-overlapping sets $\mathcal{S}^{\textit{Non}}$ and $\mathcal{S}^{\textit{Key}}$, indicating whether the starting patch $i$ contains SIFT keypoints.
By the different identities (`\textit{Non-keypoint}' patch or `\textit{Keypoint}' patch) that patches belong to, there are four different combinations between the starting patch $i$ and the target patch $j$ as:
$\bullet$ \textit{KK}: `\textit{Keypoint}'$\to$ `\textit{Keypoint}', $\bullet$  \textit{KN}:`\textit{Keypoint}'$\to$ `\textit{Non-keypoint}', $\bullet$  \textit{NK}:`\textit{Non-keypoint}'$\to$ `\textit{Keypoint}', and  $\bullet$ \textit{NN}:`\textit{Non-keypoint}'$\to$ `\textit{Non-keypoint}'.

Take the $\bullet$ \textit{KK}: `\textit{Keypoint}'$\to$ `\textit{Keypoint}' as an example, the target of patch interrelation analysis under this combination is to find `\textit{Starting from a keypoint patch, whether its attention distribution will present some special interrelation patterns towards other keypoint patches?}'
The global patch interrelation analysis under different combinations are:
\begin{equation}
\label{eq:global patch interrelation analysis}
\begin{split}
    &\theta_{\textit{KK}} = \frac{\sum\limits_{i\in \mathcal{S}^{\textit{Key}}} \theta_i}{|\mathcal{S}^{\textit{Key}}|} ; \quad \quad \quad \theta_{\textit{NK}} = \frac{\sum\limits_{i\in \mathcal{S}^{\textit{Non}}} \theta_i}{|\mathcal{S}^{\textit{Non}}|} ; \\
    &\theta_{\textit{KN}} = \frac{\sum\limits_{i\in \mathcal{S}^{\textit{Key}}} (1-\theta_i)}{|\mathcal{S}^{\textit{Key}}|} ; \quad \theta_{\textit{NN}} = \frac{\sum\limits_{i\in \mathcal{S}^{\textit{Non}}} (1-\theta_i)}{|\mathcal{S}^{\textit{Non}}|} .
\end{split}
\end{equation}


%


\section{Hidden Patterns in Patch Interrelation}
\label{Hidden Patterns in Patch Interrelation}

The global patch interrelation analysis on the entire image provides a comprehensive and novel understanding about how the ViT model utilizes patch interrelation in different stages during the model forward inference process.

\subsection{Attention Patterns among Patches}

\begin{figure}[t]
    \centering
    \includegraphics[width=0.48\textwidth]{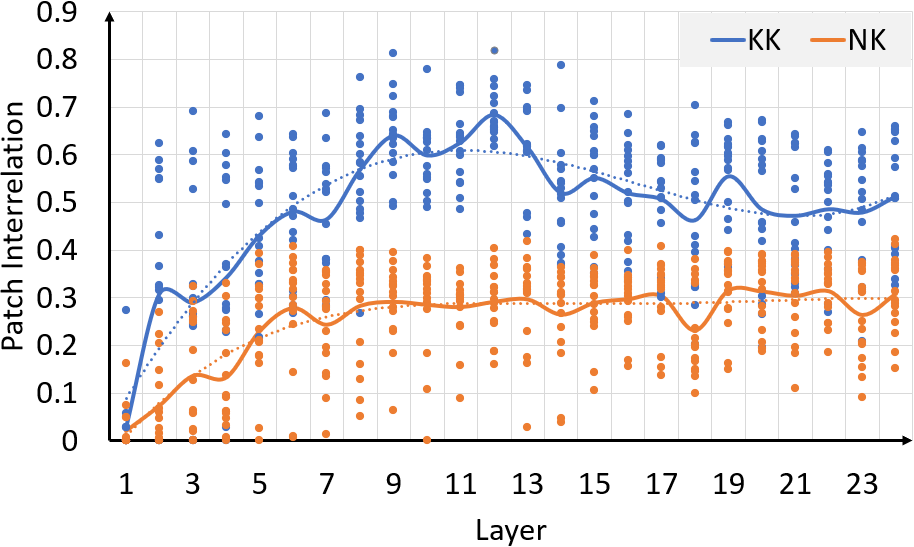}
    \caption{The global patch interrelation analysis on all $24$ transformer layers, where each point corresponds to the patch interrelation analysis $\theta_i$ on a particular SA head. The solid line is the average of all $16$ SA heads, and the dotted line is its trend line.}
    \label{fig:attention pattern}
\end{figure}


We first perform the global patch interrelation analysis on all $16$ SA heads throughout all $24$ transformer layers, where the results are demonstrated in Fig.\ref{fig:attention pattern}. 
The observation is interesting: starting from keypoint patches $\in \mathcal{S}^{\textit{Key}}$, their attention distributions are discriminative to different target patches, which are highly biased towards other keypoint patches $\in \mathcal{S}^{\textit{Key}}$.
We can see from the `\textit{KK}' case (blue line) in Fig.\ref{fig:attention pattern}, the interrelation between keypoint patches has a rapid increasing during the model forward inference process, and presents an obvious gap with the `\textit{NK}' case (orange line).
This discriminative bias of ViT model prove our conjectures from pre-experiment results on image recovery, which indicates that:
\textbf{1)} There are high semantic interrelation patterns among keypoints patches. 
\textbf{2)} The MSA mechanism of ViT model mainly utilize the keypoint patch interrelation during its forward inference process.
The analysis results on `\textit{KN}' and `\textit{NN}' cases are omitted here but attached later in Fig.\ref{fig:Three stages summary} for different stages summary, since they have a clear mathematical $1-\theta$ complementary relationship with their respective counterparts (for example, `\textit{KN}' $= 1 -$ `\textit{KK}'), as shown in Eq.(\ref{eq:global patch interrelation analysis}).

Moreover, except for the obvious gap between two cases, we can also see a dynamic change of the `\textit{KK}' case (blue line) in different stages of model forward inference process. This phenomenon indicates that the ViT model has a different semantic understanding of patch interrelation in different stages, which will be elaborated later in Sec.\ref{subsec:Capture, Focus and Understand in ViT}.


\subsection{Focused or Broad? --- Attention Focus Index}

\begin{figure}[t]
    \centering
    \includegraphics[width=0.48\textwidth]{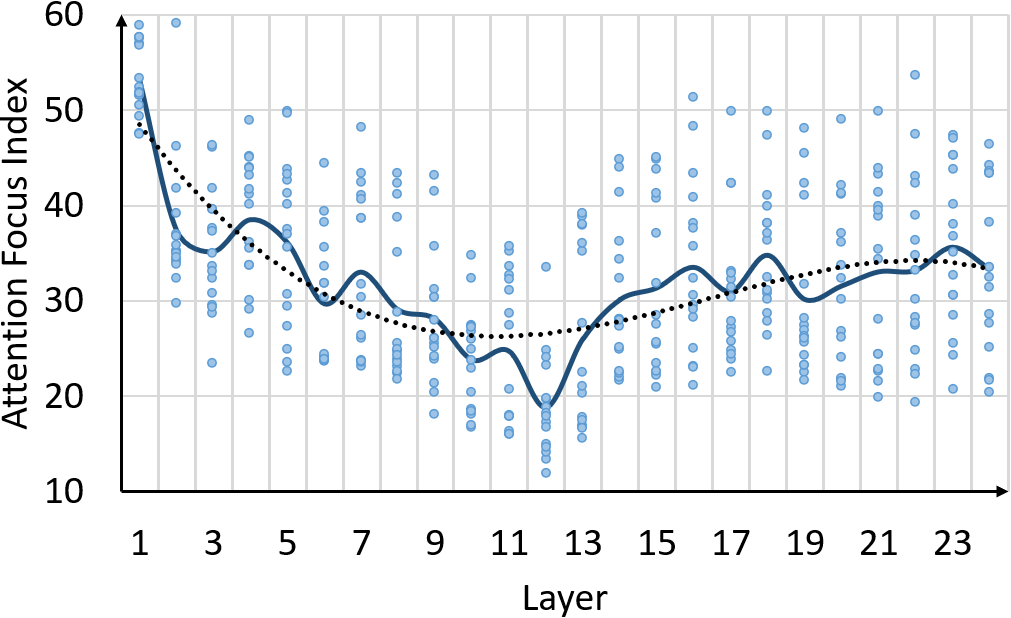}
    \caption{Attention focus index of all $16$ SA heads (blue points) in each layer. The bold line shows their average in each layer and the dash line is its trend-line. }
    \label{fig:attention focus index}
\end{figure}

Except for distribution characteristics reflected by above attention patterns, we also measure the attention focus level of ViT model, i.e., whether SA heads will focus on a few patches or attend broadly on many different patches. We use the average information entropy of each head’s attention distribution as our metric, which is called attention focus index.
For a $16\times 16$ attention matrix $\bm{\alpha}_i = [\alpha_{ij}]$ of patch $i$ generated by one SA head, we first transfer all attention value to the $[0, 1]$ interval by Min-Max normalization, and then calculate its information entropy $\delta_i$ as:
\begin{equation}
\label{eq:attention focus index}
    \delta_i =  -\sum_{j=1}^{16 \times 16} \log p\left( \frac{\alpha_{ij}- \min(\bm{\alpha}_i)}{\max(\bm{\alpha}_i) - \min(\bm{\alpha}_i)} \right).
\end{equation}
Finally, by averaging all information entropy $\delta_i$ of total $256$ different starting patch $i$ on the same SA head, we can obtain the attention focus index of that specific SA head. We show the dynamic change of attention focus index on all SA heads during the model forward inference process in Fig.\ref{fig:attention focus index}.

Intuitively, a larger attention focus index indicates that the model only attend to fewer patches, vice versa. Therefore, we can find the attention focus index has an obvious change that the model rapidly focuses to only a few patches in former layers, but back to broad attention again in later layers.
However, the implied focus patterns are highly diverse in different stages and we will provide a full analysis later in Sec.\ref{subsec:Capture, Focus and Understand in ViT} to explain them.

\subsection{Retrieval, Capture, and Coach Stages in Vision Transformer}
\label{subsec:Capture, Focus and Understand in ViT}

  

\begin{figure}[t]
  \centering
  \begin{subfigure}{0.48\textwidth}
    \includegraphics[width=\linewidth]{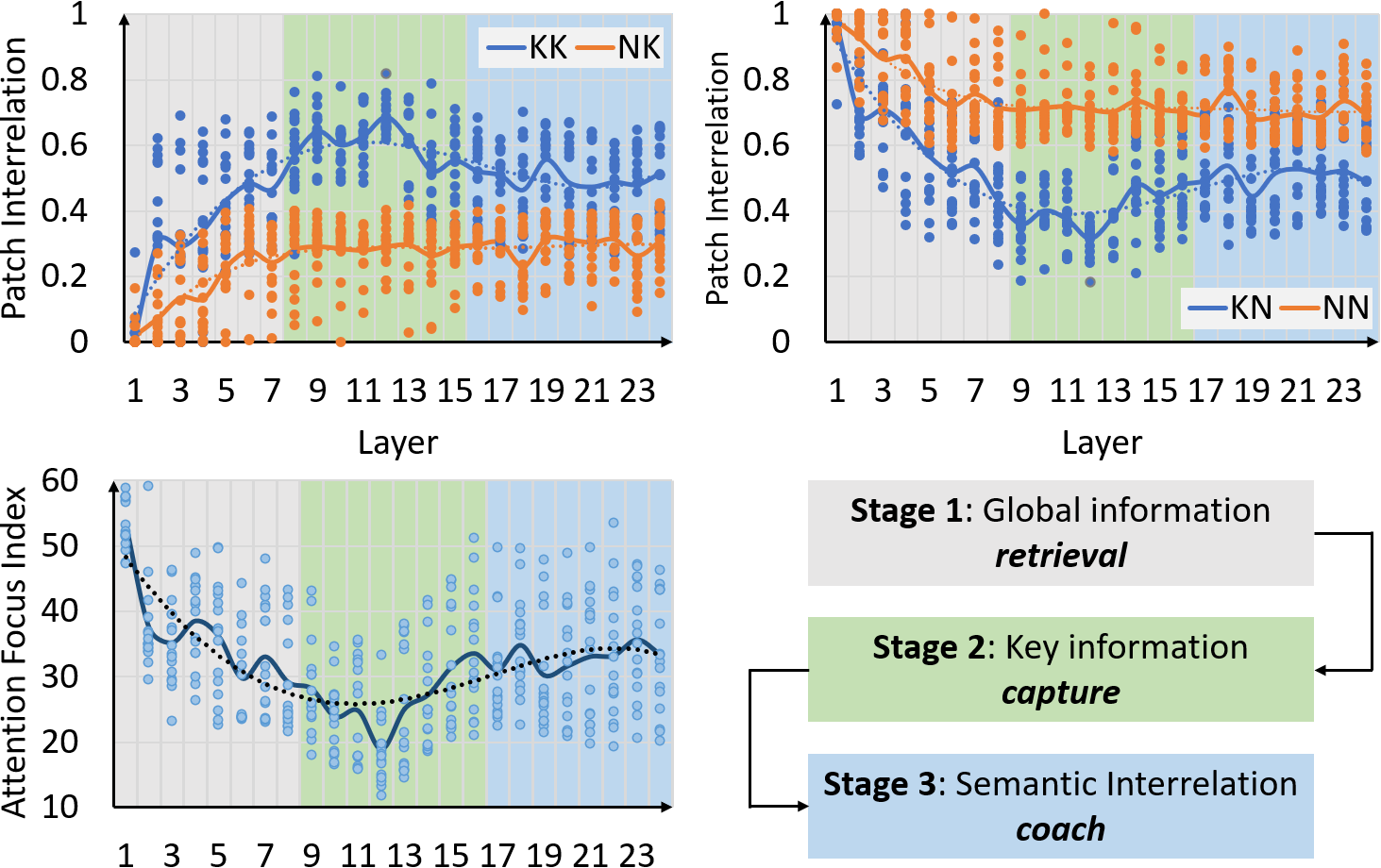}
    \caption{Three different stages identified by our quantitative analysis in model forward inference process (indicated by different background colors).}
    \label{fig:Three stages summary}
  \end{subfigure}
  \begin{subfigure}{0.48\textwidth}
    \includegraphics[width=\linewidth]{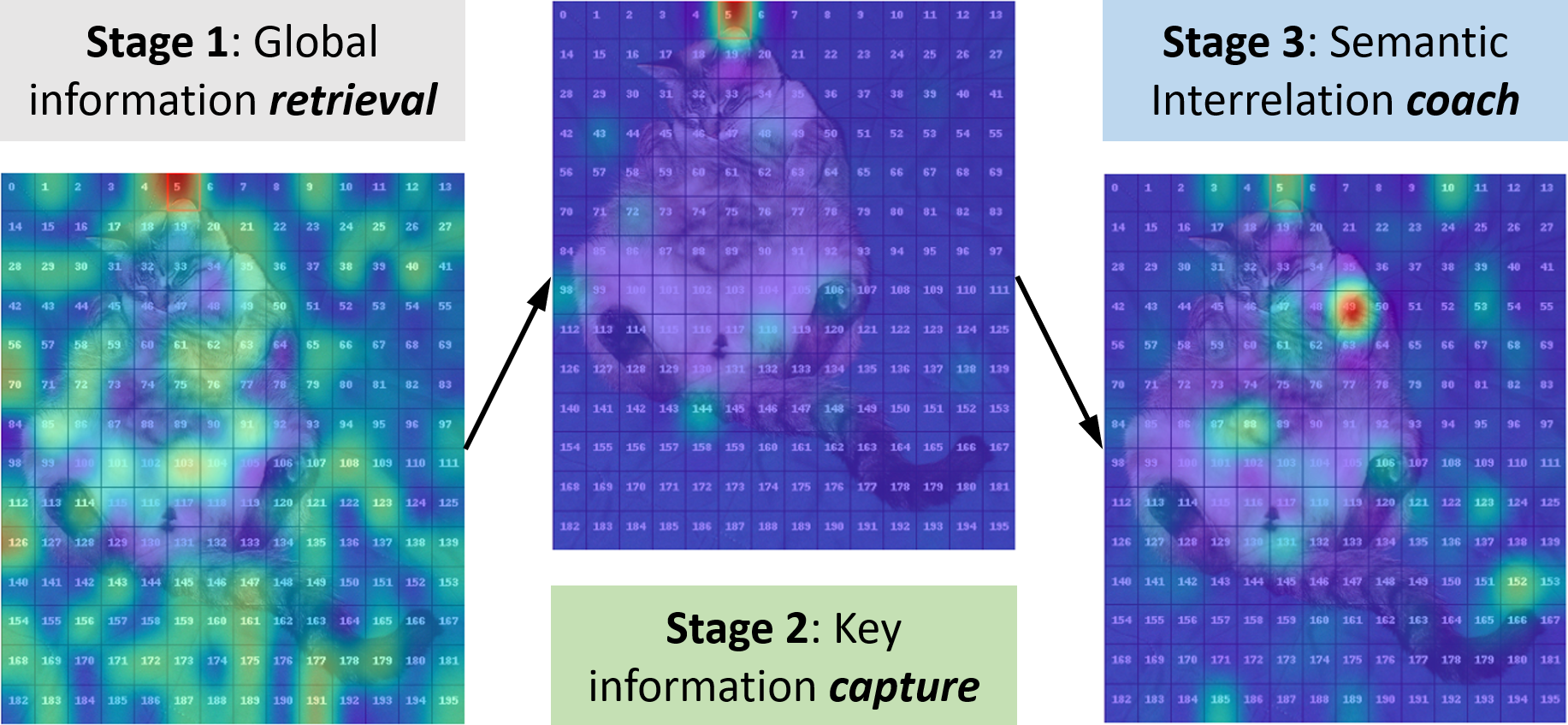}
    \caption{Visualized attention distribution heatmaps in  three different stages. }
    \label{fig:attention heatmap samples}
  \end{subfigure}
  \caption{ Three different stages (i.e., Retrieval, Capture, and Coach Stages) in model forward inference process and their visualized heatmap samples of attention distributions.}
  \label{fig:Three stages summary and visualizaiton}
  \vspace{-10pt}
\end{figure}



Combining all analysis results in previous sections, we can obviously see three different stages during the model forward inference process, which are $\bullet$ \textbf{Stage 1}: Global information \textit{Retrieval}, $\bullet$ \textbf{Stage 2}: Key information \textit{Capture} and $\bullet$ \textbf{Stage 3}: Semantic Interrelation \textit{Coach}, as shown in Fig.\ref{fig:Three stages summary} with different background colors. 
Next, with the visualized attention heatmaps in Fig.\ref{fig:attention heatmap samples}, we will provide a detailed analysis to explain the differences of three stages and how the patch interrelation is used in diverse ways.

First, in $\bullet$ \textbf{Stage 1}: Global information \textit{Retrieval}, the `\textit{KK}' and `\textit{KN}' patch interrelation are both very low (only have a small gap), which means that no matter what identity the starting patch is, the ViT model treats all target patches equally. 
Besides, the high attention focus index at this stage also indicates that the attention of ViT model is broadly distributed over all patches, as shown in the heatmap sample of Fig.\ref{fig:attention heatmap samples}. 
Therefore, the function of ViT model at this stage is mainly to retrieve all patches to integrate global information of the image, and thereby make preparations for the next stage.

Next, in $\bullet$ \textbf{Stage 2}: Key information \textit{Capture}, the `\textit{KK}' patch interrelation is rapidly increasing and opening a huge gap with the `\textit{KN}', which means that the ViT model is biased in this stage. With the help of integrated global information in last stage, it concentrates on capturing the strong interrelation between those keypoint patches now.
Meanwhile, the rapidly dropping attention focus index also suggests that the attention of ViT model is highly focused, as shown in the second sample of Fig.~\ref{fig:attention heatmap samples}.
Thus, the ViT model in this stage is to capture the key patch interrelation and refine the critical information that can help to truly understand the image in later stage.

Finally, in $\bullet$ \textbf{Stage 3}: Semantic Interrelation \textit{Coach}, we can observe a slight decrease in the `\textit{KK}' patch interrelation, along with a slight increase in attention focus index. These observations means that the ViT model reallocates a small portion of its attention back to other Non-keypoint patches, and thus results in the rebound of attention focus index.
However, different from the global information retrieval in first stage, this attention reallocation is deliberately operated by the ViT model, whose target is to establish the semantic interrelation coachings among patches with different identities. 
Although those Non-keypoint patches contain only little information, but they aren't completely useless for image understanding, since the sharp contrast between them and keypoint patches can further help the model to distinguish the true objective in the image.
The third sample of Fig.\ref{fig:attention heatmap samples} also further reflects our conclusion at this stage.


\section{Applications in Practice}

The previous patch interrelation analysis not only uncovers the hidden working mechanism of MSA in ViT model, but also provide further guidance to both model training and inference process. In this section, we list some of interesting applications in practice that can utilize the advantages of patch interrelation.

\subsection{Spurious Correlation Discovery}


Some previous studies \cite{singla2021understanding,eyuboglu2022domino} have shown that traditional image classification tasks suffer from model performance degradation due to the spurious features hidden in images. For example, blue sky background is likely to co-occur in many bird images. However, the absence of blue sky background from a bird image should not result in misclassification.
Different from their single-point spurious feature discovery, our patch interrelation can analyze the mutual attention distribution of different patch combinations, and then identify the true dominant feature correlation.
For example, a keypoint patch, that assign most of its attention to those Non-keypoint patch, will be regarded as spurious correlation, even if this patch contains many keypoints. 

\begin{figure}[t]
  \centering
  \begin{subfigure}{0.48\textwidth}
    \includegraphics[width=\linewidth]{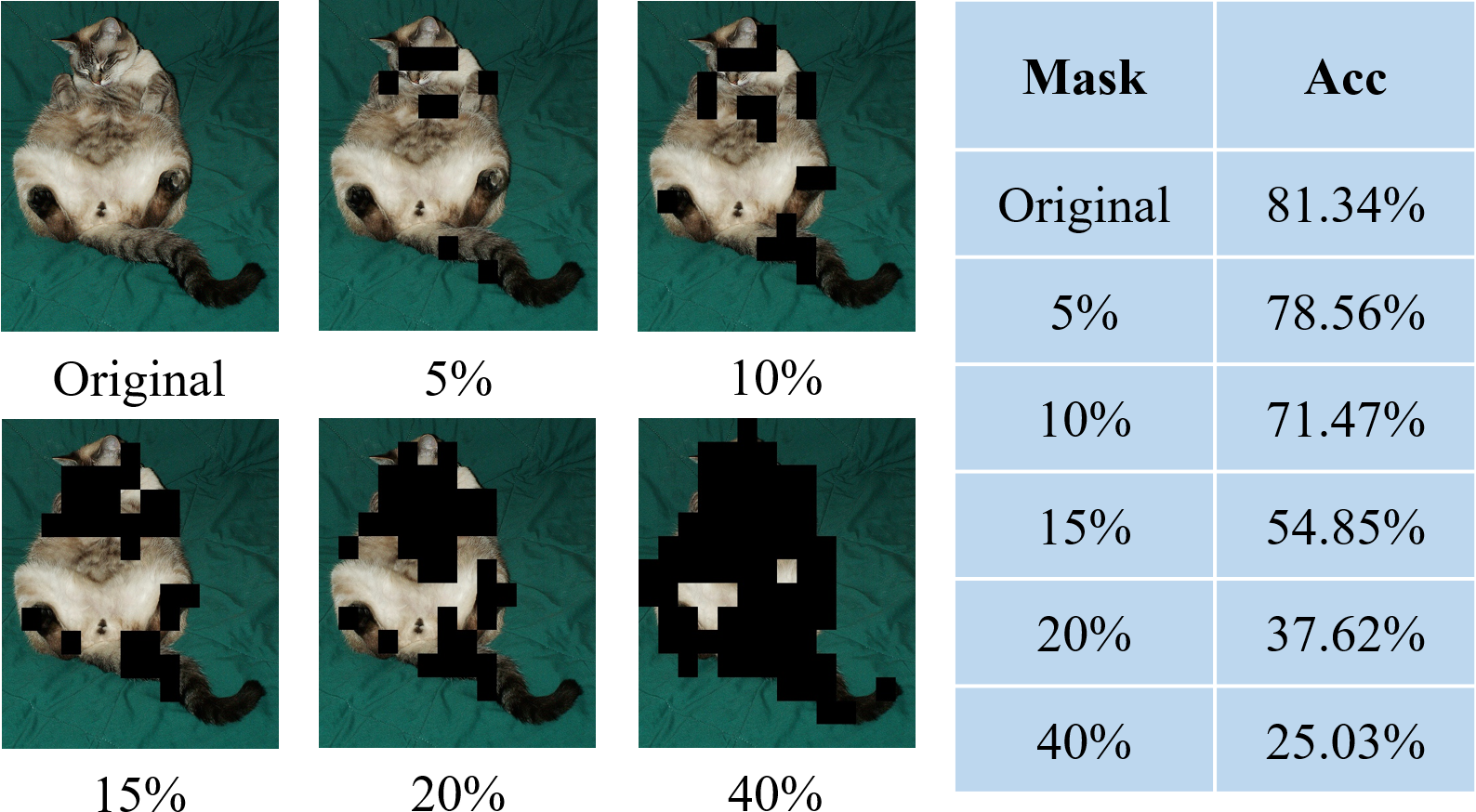}
    \caption{Visualization of \textbf{Top} patch masking under different mask ratios and the change of model accuracy }
    \label{fig:top mask}
  \end{subfigure}
  \begin{subfigure}{0.48\textwidth}
    \includegraphics[width=\linewidth]{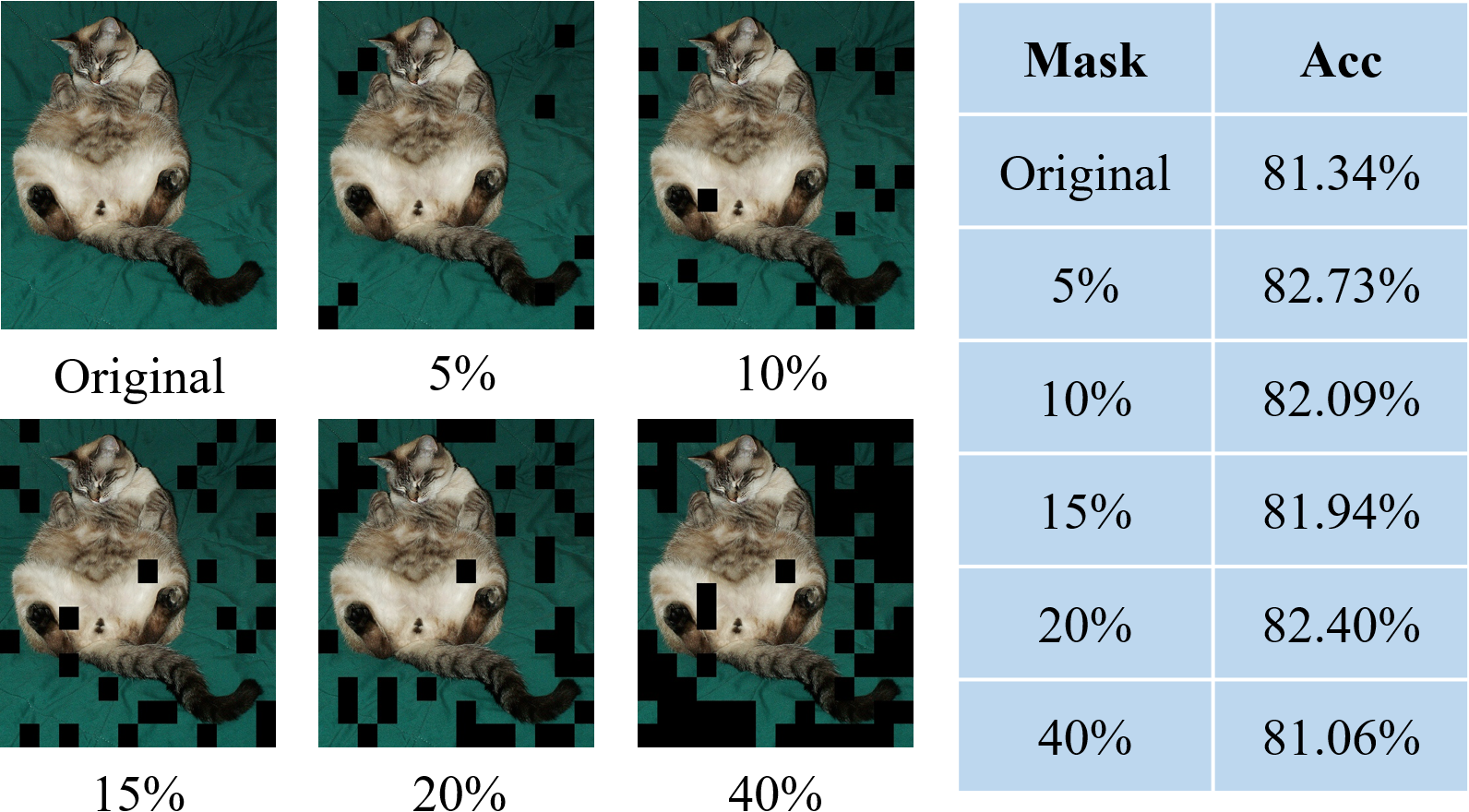}
    \caption{Visualization of \textbf{Bottom} patch masking under different mask ratios and the change of model accuracy.}
    \label{fig:bottom mask}
  \end{subfigure}
  \caption{Spurious Correlation Discovery. The \textbf{Top} case shows the importance of keypoint patch interrelation for image understanding. The \textbf{Bottom} case shows we can find the useless or even negative patch interrelation and serve as a \textit{Prompt} to further improve model inference performance. The accuracy is tested on ILSVRC2012 (ImageNet 1K) with pretrained ViT-B/16 model.}
  \label{fig:Spurious Correlation Discovery}
  \vspace{-10pt}
\end{figure}

To verify the effectiveness of our spurious correlation discovery, we design a special patch mask mechanism during the model inference process to understand what is the impact of spurious Correlation.
According to Sec.\ref{subsec:Measurement of Patch Interrelation}, we know that $\theta_i$ shows the attention level to other keypoint patches starting from patch $i$ on a specific SA head. Since we have total $m=16$ SA heads in each layer of our ViT model, we can get the average by $\bm{\theta}_i=\frac{1}{16}\sum_{m=1}^{16}\theta_i$, which indicates the average attention level of patch $i$ to other keypoint patches. In other words, each patch $i$ will have a score $\bm{\theta}_i$ that reflects its importance level of patch interrelation.
Next, we use $\bm{\theta}_i$ to implement our patch mask mechanism from two different aspects: \textbf{Top} and \textbf{Bottom}. In the Top patch mask case, we will mask a subset of patches based on their score $\bm{\theta}_i$ from highest to lowest, where $r$ is the mask ratio (vice versa for the bottom case). We set different mask ratio from $r=5\% \to 40\%$, and visualize all mask results as shown in Fig.\ref{fig:Spurious Correlation Discovery}.

Intuitively, in the \textbf{Top} case, all masked patches are the most important keypoint patches that contain the critical semantic information for understanding the image. 
Therefore, we can clearly observe a rapid image classification accuracy drop as the increasing of mask ratio in the table of Fig.\ref{fig:top mask}. 
Besides, when the mask ratio $r$ is relatively small (such as $5\%$ and $10\%$), the ViT model can still use the patch interrelation from the remaining to understand the missing content laterally, which results in a slow performance degradation. 
Once the mask ratio $r$ is too large that the remaining patches cannot provide enough semantic information for understanding, the model accuracy will quickly broke down, and the crash point is between $10\%$ and $20\%$. .


\textbf{Mask as Prompt}. In contrast to the straightforward results in previous Top case, we have some more interesting finding in the bottom case, which can further improve the model inference performance. 
Since patches in bottom case are masked according to the lowest ordering of $\bm{\theta}_i$ value, which means that they can provide very little semantic information and patch interrelation to understand the image. 
What's worse, there is usually some redundant background in the image that has no relevance to the real objective, they may even induce negative information to hinder the image understanding. 
Removing these negative information from the original image will potentially drive the model to focus more on the true objective and learn the real patch interrelation hidden in the image.
This concept is borrowed from \textit{Prompt Learning} \cite{zhou2022learning,jia2022visual}, which only introduce small number of extra trainable parameters (called Prompt) in the original inputs to adapt different downstream tasks, thereby improving the model inference performance. 
Follow the similar insights, the patch mask mechanism is served as the Prompt in our case.

From the experiment results in the table of Fig.\ref{fig:bottom mask}, We can see that the model classification accuracy has a slight increasing when the mask ratio is relatively small (see 5\% and 10\% cases), which means these masked patches will provide negative information and interrelation to hinder image understanding before. This phenomenon is also consistent with our concept of "Mask as Prompt".
Besides, even if the mask ratio $r$ is later increased to $40\%$, there is no obvious accuracy drop compared to the original image, which empirically proves the effectiveness of our patch interrelation analysis to  find the hidden spurious correlation in the image.


\subsection{Guided Mask for Model Pre-training}


Except the previous applications for model inference, Our patch interrelation analysis can also be utilized in the model pre-training process to speed up the training efficiency. 
The existing model pre-training of MAE encoder is based on a random mask mechanism \cite{he2022masked}, which only randomly samples a patch subset to pre-train ViT model. When facing the agnostic patch interrelation, random mask is a straightforward approach to effectively ensure the model pre-training performance. As long as the random sampling time (training time) is large enough to cover different patches combinations, random mask can guarantee ViT model to understand the hidden semantic information in images (i.e., various patch interrelation in our paper).

However, according to the studies in Curriculum Learning (CL) \cite{bengio2009curriculum}, the model training process is similar with human learning process, it is advocated to let the model training start from easy data samples first and gradually progress to complex data samples and knowledge.
Through a series of well-designed learning tasks, the CL strategy can accelerate the model training process to reach the same model performance with less training iterations.
Therefore, based on the insights from CL, we propose to gradually increase the dataset difficulty during ViT model pre-training process, thus to guide and accelerate the model learning. 
According to previous $\bm{\theta}_i$ derived from our patch interrelation analysis, each patch $i$ will have a score indicates its importance of semantic understanding to other keypoint patches, and we can use it to design our datasets with different difficulty.

Assume the mask ratio $r$ is a constant, such as $r=50\%$. For random mask, it just needs to randomly mask $50\%$ patches and feed the remaining patches into model for pre-training, where the percentage of Keypoint patches and Non-keypoint patches in masked patches are also random. If the percentage of Keypoint patches in all masked patches is denoted by $\beta$, we can now adjust the dataset difficulty by setting different $\beta$ value. First, we calculate the masked Keypoint/Non-keypoint patch number by: "total patch number $\times r \times \beta/(1-\beta)$ ", respectively.
Then, we shuffle the two lists of all Keypoint/Non-keypoint patches and sample the corresponding number of patches from the top of Keypoint patch list and the bottom of Non-keypoint patch list, where these sampled patches are masked.
Obviously, a larger $\beta$ value means more Keypoint patches are masked, which indicates higher dataset difficulty.
When $\beta=0\%$, all masked patches are Non-keypoint patches (lowest dataset difficulty), vice versa for $\beta=100\%$. 

\begin{figure}[t]
    \centering
    \includegraphics[width=0.48\textwidth]{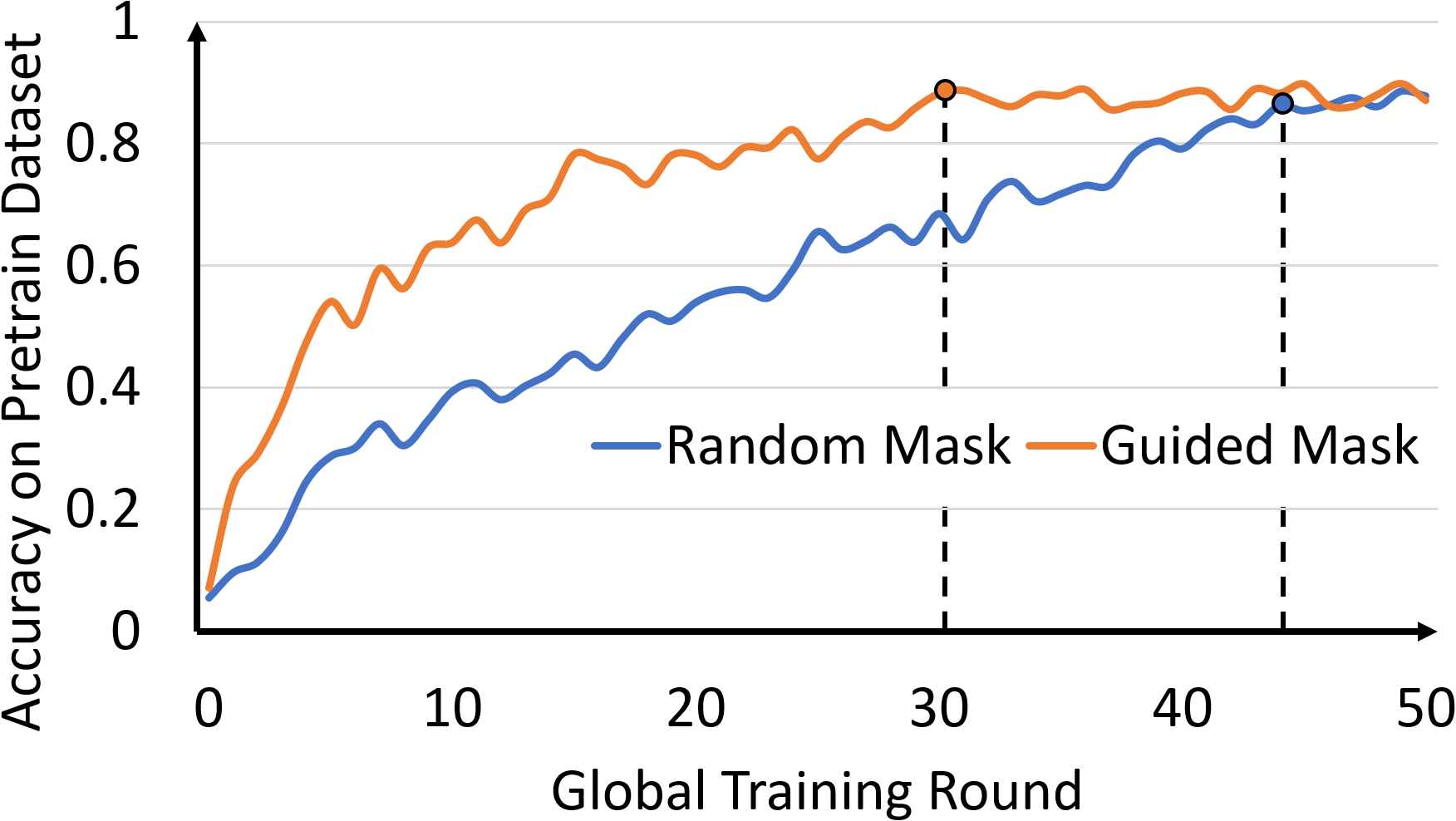}
    \caption{The ViT-Base/16 model pre-training on ImageNette with different mask mechanisms, where our guided mask can  accelerate pre-training process to reach convergence with less rounds.}
    \label{fig:Guided mask}
    \vspace{-10pt}
\end{figure}

In our experiment, we design $5$ datasets with increasing difficulty from $\beta = 10\%$ to $\beta = 50\%$, and gradually increase the dataset difficulty every $10$ rounds.
The experiment configurations are listed as below: model (ViT-Base/16), Pre-train dataset (ImageNette \footnote{https://www.tensorflow.org/datasets/catalog/imagenette}), GPU (one RTX 3090). We use the classification accuracy on the pretrain dataset to evaluate the model pre-training performance.
It is clear from Fig.\ref{fig:Guided mask} that our guided mask can reach convergence point at around $30$-th training round, while the random mask needs to take about $45$ training rounds.

% ...


\section{Related Work}
\label{sec:related work}

\subsection{Transformer-based models for vision tasks}
\label{subsec:Transformer-based models for vision tasks}
Different from the traditional CNN-based models that use convolution kernel to extract features from the original image for later tasks, the Transformer-based model architecture mainly utilize the self-attention mechanism to capture the long-term dependencies between different parts of the image. 
Vision Transformer (ViT) model is the first work that applies MSA mechanism on image, by splitting it into a series of non-overlapping patches as the input of Transformer model \cite{dosovitskiy2021image}. The flexibility and effectiveness of ViT make it becomes the ubiquitous landmark technique in computer vision field.
Extensive subsequent studies based on ViT are proposed to design more efficient MSA mechanisms, including implementing deeper ViT model with diverse regenerated attention map \cite{zhou2021deepvit}, constructing hierarchical ViT structure to apply the shift window scheme and limiting the focus of self-attention mechanism \cite{liu2021swin},
and multi-scale ViT structure with attention mechanism of Vision Long former \cite{zhang2021multi}
However, despite the promising performance, all these studies have ignored a fundamental problem, i.e., \textit{how the MSA mechanism works in vision?}

\subsection{Interpretability of Vision Transformer}
\label{subsec:Interpretability of Transformer}

When ViT first demonstrate its powerful performance to outperform previous CNN-based baselines \cite{dosovitskiy2021image}, its unique model structure has attracted extensive focus from researchers to understand its interpretability from different aspects, including: observing the attention map of Transformer outputs \cite{li2020does,chefer2021transformer}, computing the relevancy of different attention heads in Transformer networks \cite{clark2019does,voita2019analyzing}. 
However, their explanations all come from the empirical observations based on Transformer network outputs or layer similarity, without any interpretable analysis backed by a solid theoretical foundation. Different from them, we try to understand the interpretability of ViT from a semantic perspective similar to NLP, and prove that ``\textit{the working principle of MSA mechanism in ViT is consistent with human semantic understanding on images}''.

\section{Discussion \& Conclusion}
In this paper, we find that the low semantic information concentration of pixels in images is a critical challenge that hiders the interpretability of ViT model. To fill this missing gap, we introduce the concept of SIFT to map the low-level redundant image space to a mid-level semantic feature space, which can automatically annotate the invariant
feature keypoints in the image, and provide a theoretical cross-correlation for measuring the semantic information level.
Then, we design a weighted quantitative analysis to explore the hidden attention patterns in the MSA mechanism, whose interpretability is consistent with the semantic understanding based on the SIFT keypoints.
Finally, we further derive a series of well-suited practical applications that benefit a lot from our interpretable analysis, including: 1) discovering spurious correlations and prompting input during model inference to improve performance, and 2)  guiding model pre-training to accelerate convergence.

{\small
\bibliographystyle{ieee_fullname}
\bibliography{egbib}
}

\end{document}